\documentclass[conference]{IEEEtran}
\IEEEoverridecommandlockouts

\usepackage[hyphens]{url}
\usepackage{hyperref}
\hypersetup{colorlinks=true,allcolors=black}

\usepackage{algorithm}
\usepackage{algpseudocode}
\usepackage{cite}
\usepackage{amsmath,amssymb,amsfonts}
\usepackage{graphicx}
\usepackage{textcomp}
\usepackage{multirow}
\usepackage{pgfplots}
\usepackage{xcolor}
\usepackage{subcaption}
\usepackage[T1]{fontenc}

\usepackage{hyperref}
\usepackage{url}
\def\BibTeX{{\rm B\kern-.05em{\sc i\kern-.025em b}\kern-.08em
    T\kern-.1667em\lower.7ex\hbox{E}\kern-.125emX}}
    
\begin{document}

\title{Auditing Approximate Machine Unlearning for Differentially Private Models}

\author{
\IEEEauthorblockN{Yuechun Gu \quad Jiajie He \quad Keke Chen}
\IEEEauthorblockA{\textit{University of Maryland, Baltimore County}\\
Baltimore, MD, USA\\
\texttt{\{ygu2,jiajieh1,kekechen\}@umbc.edu}}
}

\maketitle

\begin{abstract}
Approximate machine unlearning aims to remove the effect of specific data from trained models to ensure individuals' privacy. Existing methods focus on the removed records and assume the retained ones are unaffected. However, recent studies on the \emph{privacy onion effect} indicate this assumption might be incorrect. Especially when the model is differentially private, no study has explored whether the retained ones still meet the differential privacy (DP) criterion under existing machine unlearning methods. This paper takes a holistic approach to auditing both unlearned and retained samples' privacy risks after applying approximate unlearning algorithms. We propose the privacy criteria for unlearned and retained samples, respectively, based on the perspectives of DP and membership inference attacks (MIAs). To make the auditing process more practical, we also develop an efficient MIA, A-LiRA, utilizing data augmentation to reduce the cost of shadow model training. Our experimental findings indicate that existing approximate machine unlearning algorithms may inadvertently compromise the privacy of retained samples for differentially private models, and we need differentially private unlearning algorithms. For reproducibility, we have pubished our code:\footnote{ \url{https://anonymous.4open.science/r/Auditing-machine-unlearning-CB10/README.md}}
 \end{abstract}

\begin{IEEEkeywords}
component, formatting, style, styling, insert.
\end{IEEEkeywords}

\section{Introduction}
The \textit{``right to be forgotten''} allows data contributors to request the deletion of their data from an organization's records. Recent regulations, including the General Data Protection Regulation (GDPR) in the European Union \cite{gdpr2016}, the California Consumer Privacy Act (CCPA) in the U.S. \cite{ccpa2018}, and Canada's Personal Information Protection and Electronic Documents Act (PIPEDA) \cite{pipeda2000}, have solidified this right. Consumers are also actively exercising this privacy right. For example, Google received over 3.2 million requests to remove specific URLs from search results over five years, demonstrating the importance and scale of this issue.

For machine learning models, exercising the right to be forgotten entails not only removing the affected training data samples but also ensuring models are retrained with the updated dataset.  However, with the rise of large models, frequent model retraining brings unbearable costs to model owners. Thus, the approaches to avoid high complete retraining costs, known as \textit{machine unlearning}, have emerged as a promising solution. Most recent machine unlearning methods \cite{golatkar2020eternal,foster2024fast,fan2024salun} claim that they can effectively achieve privacy protection. In particular, it's well-believed that exact unlearning methods \cite{xu2024machine} can surely achieve the privacy-protection goal as the target samples are exactly removed during the unlearning process, while approximate unlearning may leave residual information about the removed data \cite{chen2021machine}. 

While the focus remains on the unlearned samples, no unlearning method cares about the privacy of retained samples. It becomes critical to examine retrained samples when the original machine learning model is differentially private \cite{abadi2016deep}. We must also maintain the level of privacy protection for the retained samples, i.e., whether the privacy budget, $\epsilon$, is still met. Some recent studies indicate this omission might lead to severe privacy damage. Carlini et al. \cite{carlini2022Onion} show that a sample's privacy is a \textit{relative} notion, meaning that removing some samples could inadvertently increase the privacy risks of other samples in the retained data. It is consistent with the understanding of anonymization by \emph{blending in the crowd} -- a thinning crowd reduces protection to the crowd members. No study has shown how this \emph{privacy onion effect} may affect machine unlearning results.

\textbf{Scope of Research}. We take a holistic approach to audit the effect of approximate machine unlearning algorithms on both unlearned and retained samples. Specifically, we redefine the criteria for successful machine unlearning in privacy protection\footnote{Machine unlearning can also be used to remove the effect of malicious or erroneous samples \cite{xu2024machine}, which may have different measuring criteria.} for differentially private models: (1) the privacy risk of unlearned samples should be reduced to \emph{safe level} after unlearning, and (2) the privacy risk of retained samples should remain below the privacy budget. 

Recent studies on auditing differentially private machine learning algorithms provide us with ideas to formally analyze and define the above two goals. The auditing mechanism uses the theorem that if a ($\epsilon,\delta$)-differentially-private machine learning algorithm is correctly implemented, the most powerful membership inference attack (MIA) on \emph{every sample} must satisfy $\ln(\text{TPR}/\text{FPR}) < \epsilon$, where TPR and FPR are the true positive and false positive rates of the membership inference test, respectively \cite{dwork2014algorithmic,tramer2022debugging,jagielski2020auditing}. 

With a sample-level privacy risk measure, we can formally define the new criteria for machine unlearning methods. Specifically, we define that unlearned samples need to meet an arbitrarily small privacy budget, e.g., a small value $t$, $t=0.01$, so that a powerful MIA on an unlearned sample meets $\ln(\text{TPR}/\text{FPR}) < t$. Meanwhile, the retained samples should still meet $\ln(\text{TPR}/\text{FPR}) < \epsilon$. We will use these criteria to examine the existing unlearning algorithms. 

To make auditing effective, the MIA test must be effective, e.g., $\ln(\text{TPR}/\text{FPR})$ is as close to its theoretical upper bound as possible. So far, the LiRA method is considered the best MIA method that can measure an individual sample's $\ln(\text{TPR}/\text{FPR})$ more accurately than other methods\cite{carlini2022Lira,zarifzadeh2024low}. However, a major challenge in applying the LiRA method is its high computational cost. Online-LiRA requires 4.8 GPU hours per sample to generate the TPR/FPR measure. While offline-LiRA is more efficient due to its one-sided hypothesis testing, its performance is also reduced. In this paper, we introduce a more efficient augmentation-based likelihood ratio attack (A-LiRA), which achieves comparable quality to online-LiRA with an 88.3\% reduction in time cost. A-LiRA is inspired by \cite{mattern2023membership}. Instead of training multiple shadow models as in online-LiRA, A-LiRA approximates the two probability distributions -- when the model is trained with or without the target sample -- using augmented data and with the assumption of normal distributions for in- and out-training measures, separately. This significantly reduces the time cost while providing a good approximation of the probability distributions, making A-LiRA both effective and efficient.

With A-LiRA, we have conducted extensive evaluations to show that many recent approximate machine unlearning methods do not satisfactorily protect the privacy of unlearned samples and, in some cases, increase the privacy risk of retained samples. Thus, new unlearning methods are desired to meet the proposed privacy measures. 

In summary, our contributions are as follows:

\begin{itemize}

    \item We re-formulate the criteria for privacy protection of machine unlearning methods for differentially private models. 

    \item We introduce a novel augmentation-based likelihood-ratio attack to estimate the privacy risk of samples, which enables efficient auditing of unlearning algorithms.

    \item We show that most approximate machine unlearning methods do not satisfactorily protect privacy for both the unlearned and retained samples.

\end{itemize}

The remaining sections are organized as follows: Related works (Section~\ref{sec:relatedworks}), preliminaries (Section~\ref{sec:prelimineries}), threat model (Section~\ref{sec:threat-modeling}), holistic auditing machine unlearning (Section~\ref{sec:standard}), and experiments (Section~\ref{sec:Exp}).

\section{Related Works}

\label{sec:relatedworks}

\textbf{Machine unlearning.} Machine unlearning enforces the ``right to be forgotten" and is categorized into exact and approximate methods. Exact unlearning retrains models without the forgotten data. A notable example is SISA by \cite{bourtoule2021machine}, which partitions training data into micro-shards and trains small models on each. To unlearn data, only the relevant micro-models are retrained, improving efficiency over full retraining. Approximate unlearning, on the other hand, employs techniques such as weight manipulation and fine-tuning to make the model ``forget" the data without retraining it entirely. For instance, \cite{golatkar2020eternal} modifies the weights to make probing functions of the weights indistinguishable from those applied to a neural network trained without the data being forgotten. \cite{fan2024salun} introduces the concept of ``weight saliency'' for machine unlearning, drawing parallels with input saliency used in model explanation. This approach focuses unlearning efforts on specific model weights, improving both efficiency and effectiveness. While machine unlearning is touted as a means to protect privacy \cite{cao2015towards,nguyen2022survey,xu2024machine}, studies such as those by \cite{chen2021machine} and \cite{carlini2022Onion} indicate that unlearning may make retained samples more susceptible to MIAs. These findings highlight the necessity for robust auditing methods and standards to ensure the privacy protection efficacy of machine unlearning.

\textbf{Auditing differential privacy (DP).} Auditing DP aims to detect potential bugs when implementing DP in practice \cite{tramer2022debugging}. Current auditing methods are derived from the hypothesis interpretation of differential privacy, which emphasizes that if an algorithm is ($\epsilon$,$\delta$)-differentially-private, any distinguishing attack's $\text{TPR}/\text{FPR}$ should be smaller than or equal to $e^\epsilon$ \cite{kairouz2015composition, dwork2008differential}. As a result, Jagielski et al. use an inversion attack to estimate the population-level $\text{TPR}/\text{FPR}$ to audit DP \cite{jagielski2020auditing}. Nasr et al. designed a membership inference method to tightly audit the DP \cite{nasr2023tight}. Steinke et al. designed an efficient method to audit the DP in only one training epoch \cite{steinke2023privacy}. However, these audits are based on the population-level $\text{TPR}/\text{FPR}$, which does not align with the per-sample assumption of hypothesis interpretation of DP. Aerni et al. emphasize that sample-level $\text{TPR}/\text{FPR}$ is a more proper metric to audit the DP and estimate the privacy risks of each sample \cite{aerni2024evaluation}. In this paper, we adopt sample-level estimation of privacy risks.

\section{Prelimineries}
\label{sec:prelimineries}
In this section, we introduce the notations and basic concepts used in later sections.

\subsection{Notations}
We denote the training dataset as $D$ and the subset to be unlearned as $X$. The retained samples are represented by $D \setminus X$. The model trained on $D$ using algorithm $\mathcal{M}$ is denoted as $M_D = \mathcal{M}(D)$, while $U_{M_D, X} = \mathcal{U} (M(D), X)$ represents the unlearned model after applying the unlearning mechanism $\mathcal{U}$ on $X$ and the existing model $M_D$. Correspondingly, the complete retraining leads to a model, $M_{D \setminus X}$. For membership inference attacks (MIA), $\mathcal{A}(U, D, x)$ indicates an MIA applied to the unlearned model $U$ for a target sample $x$ with the known distribution of $D$. We use $E(M_D,x)$ denote the $\text{TPR}_x/\text{FPR}_x$ of $\mathcal{A}$ on model $M_D$.

\subsection{Differential Privacy}

Differential Privacy (DP) ensures that the inclusion or exclusion of a single data point minimally affects the output of an algorithm, protecting individual privacy. A machine learning algorithm $\mathcal{M}$ satisfies $(\epsilon, \delta)$-DP if, for any two neighboring datasets $D$ and $D'$ that differ in at most one sample, and any output set $S$:

$$
\Pr[ \mathcal{M}(D) \in S] \leq e^{\epsilon} \cdot \Pr[ \mathcal{M}(D') \in S] + \delta
$$

Here, $\epsilon$ (privacy loss) controls the strength of the guarantee, with smaller $\epsilon$ offering stronger privacy. $\delta$ allows for a small probability of privacy violation, especially for extreme outputs. Lower values of $\epsilon$ and $\delta$ indicate higher privacy protection.

\subsection{Likelihood Ratio Attack}
\label{sec:LiRA}
Likelihood ratio attacks (LiRA) were first introduced by \cite{carlini2022Lira} for machine learning models and advanced by \cite{zarifzadeh2024low}. Given a target sample $x$, the online-LiRA sets up two hypotheses:
\begin{align*}
H_0 &: \text{Target model is trained on } D, \\
H_1 &: \text{Target model is trained on } D \setminus x
\end{align*}
It trains multiple shadow models (typically 512 in total) on $D$ or $D \setminus x$ fits the logit outputs to Gaussian distributions. The likelihood ratio between the two distributions is used to reject one of the hypotheses. While effective, this method is computationally expensive, taking an average of 4.8 hours per sample in CIFAR-10 on a Titan V100 GPU, making it impractical. A more efficient variant, offline-LiRA, only estimates the distribution $\mathcal{D}(x| \mathcal{M}(D \setminus x))$ of the target model is not trained on the target model and uses the likelihood of $x$ follows the estimated distribution to reject or accept $H_1$. This one-sided approach can be batched by training shadow models on random subsets of $D$, but it performs worse than the online-LiRA in efficacy.

\subsection{Estimating Per-sample Privacy Risk}
\label{sec:estimation}
There are several metrics to estimate the per-sample privacy risk, e.g., per-sample attack success rate of LiRA \cite{carlini2022Onion} and fisher information of samples \cite{farokhi2017fisher}. We adopt the per-sample $\text{TPR}/\text{FPR}$ of MIAs as the privacy risk metric because it intrinsically relates to differential privacy and has been widely used \cite{aerni2024evaluations,nasr2023tight,steinke2024privacy,tramer2022debugging,Gu2024,he2025recps}. To estimate the privacy risk of $x_i \in X$, we train $n$ models. Each model is trained on a random half of $X$ so that each $x_i$ is used to train around $m/2$ models. Then we use an effective MIA, $\mathcal{A}$, to attack each $x_i$ on each of $n$ models. Thus, we have $n$ true memberships and predicted memberships of each $x_i$. Then we compute $\text{TPR}_{x_i}/\text{FPR}_{x_i}$ as the privacy risk of $x_i$. To estimate the privacy risks after unlearning, we simply unlearn specific samples from each of $n$ models and recompute the privacy risks of each sample. For the unlearned samples, we keep the true membership unchanged after unlearning because we want to measure how the unlearned model "forgets" the samples. Ideally, we expect a smaller $\text{TPR}/\text{FPR}$ on both unlearned and retained samples, which implies the sample becomes safer. However, we empirically find that a small fraction of both unlearned and retained samples will become reskier, which shows that the machine unlearning does not always protect the privacy of unlearned samples and can even breach the privacy of retained samples.

\section{Threat Model} 
\label{sec:threat-modeling}
Before introducing the measure for auditing machine unlearning, we outline the threat model for our proposed attack.

\textbf{Target Model and Sample.} The target model consists of an original model, $M_D$, trained on dataset $D$, and an unlearned model, $U_{M_D, X}$. According to the ideal definition of machine unlearning (\cite{xu2023machineunlearningsurvey,nguyen2022survey}), the unlearning method should be equivalent to the model retrained on the retained set $D\setminus X$, i.e., $U_{M_D, X} \approx M_{D\setminus X}$. However, approximate unlearning does not normally meet this strict condition. Thus, it is important to know how much the effect of the target samples on the model is reduced, which is challenging to evaluate.

\textbf{Attacker's Capabilities.} To assess the privacy risks associated with individual samples, we consider a worst-case adversary. The attacker has white-box access to both the original model $M_D$ and the unlearned model $U_{M_D, X}$, as well as full knowledge of the training dataset $D$ and the unlearned samples $X$. 

\textbf{Attacker's Objective.} The attacker's goal is to determine whether a sample $x \in D$ was used to train the model $M_D$, or a sample $x \in D\setminus X $ was used to train the model $U_{M_D, X}$. 

\textbf{Success of Protection.} We define the success of differential privacy protection as the attacker's MIA ability on the unlearned model for any sample $x \in D\setminus X$, denoted as $\mathcal{A}(U, D, x)$, is bounded by the differential privacy budget $\epsilon$. Meanwhile, any unlearned sample $x \in X$ should have a risk smaller than a preset value $t$, where $0 \leq t \ll \epsilon$.

\section{Auditing Machine Unlearning}
\label{sec:standard}
So far, all existing machine unlearning algorithms assume we only need to ensure privacy protection for the users who execute ``the right to be forgotten'', and the retained samples' privacy protection is fine to be omitted in unlearning. This assumption is incorrect if the original model is differentially private, where the model builder has reached an agreement of privacy guarantee, i.e., the $\epsilon$ setting, with data contributors. The ``privacy onion effect'' \cite{carlini2022Onion} has shown that the removal of some samples may affect other samples' privacy risks. Thus, we take a holistic approach to define the criteria for auditing \emph{all samples' privacy guarantees} and develop an efficient algorithm to support the auditing process.

\subsection{Auditing Criteria}
We use an effective MIA, $\mathcal{A}$, for our auditing purpose, which gives $\text{TPR}_x$ and $\text{FPR}_x$ for every sample $x$. Now we explain the auditing criteria for unlearning a differentially private model $M_D$. Note that we target unlearning for differentially private models with a privacy budget $\epsilon$. 

 \textbf{Criterion 1: For the unlearned samples $X$}, will unlearning $X$ adequately protect its privacy? Formally, let $E(M_D,x)$ denote the $\text{TPR}_x/\text{FPR}_x$ of $\mathcal{A}$ on model $M_D$. For every sample $x \in X$, we have 
 
\begin{equation*}
 E(U_{{M_D},X}, x) \approx E(M_{D\setminus X}, x), x\in X
\end{equation*}

This equation indicates that the privacy risk of unlearned samples on the unlearned model should approximate that in a retrained model. In practice, without knowing the retrained model, the following inequality \textit{must} hold for an effective unlearning algorithm:

\begin{equation}
   \label{eq:1}
 E(U_{{M_D},X}, x) < E(M_{D}, x) - t_1, x\in X
\end{equation}

where $t_1\geq 0$ is a pre-defined threshold that model builders use to achieve stricter or relaxed privacy guarantees. When $t_1 = 0$, the equation reflects the minimum requirement on the unlearning algorithm. That is, the privacy risk of unlearned samples on unlearned models should be less than that on the original model. Violating Equation~\ref{eq:1} at $t_1=0$ implies the unlearning algorithm does not achieve the goal of privacy protection for unlearned samples. 

\textbf{Criterion 2: For the retained samples in $D \setminus X$}, will unlearning $X$ increase the privacy risk of the retained data? When the original algorithm is ($\epsilon$,$\delta$)-differentially private, we have
\begin{equation*}
   E(M_D, x) \leq \epsilon, x\in D
\end{equation*}

This equation is derived from the relationship between distinguishing attacks and differential privacy as introduced in Section~\ref{sec:relatedworks}. 

Ideally, the unlearned models also have to meet the privacy guarantee

\begin{equation}
\label{eq:2}
E(U_{{M_D},X}, x) \leq \epsilon, x \in D\setminus X
\end{equation}

This implies that the privacy risk of retained samples on unlearned models should not exceed the pre-set theoretical $\epsilon$ in the differentially private $\mathcal{M}(D)$. 

We use these two criteria to determine whether a machine unlearning method is applicable to protect privacy. For a differentially private model, the unlearned model is applicable only when it satisfies both Equation~\ref{eq:1} and ~\ref{eq:2}.

\subsection{Augmentation-based Likelihood Ratio Attack (A-LiRA)}
\label{sec:A-LIRA}
To check the proposed auditing criteria, we need an accurate and efficient MIA to estimate the privacy risk at the sample level. This estimator must be theoretically sound and computationally efficient. 

The attack $\mathcal{A}$ must be powerful to maximize the $\text{TPR}/\text{FPR}$ ratio, in particular, achieving high TPR at low FPR, which is critical for security analysis \cite{kolter2006learning,kantchelian2015better,carlini2022Lira}. So far, LiRA \cite{carlini2022Lira} is the most accurate MIA algorithm in the low-FPR region, but the online LiRA is costly, while the offline LiRA sacrifices accuracy, which may not effectively audit the unlearning process. Assume $IN$ and $OUT$ represent the training dataset and the non-training dataset, respectively. The performance bottleneck of online LiRA stems from training many shadow models to distinguish how the target sample $x$ behaves in $IN$ and $OUT$ datasets. Online LiRA trains $2n$ shadow models to fit two distributions ($n=256$ typically) for $IN$ and $OUT$, respectively. It is an expensive procedure, e.g., costing about 4.8 GPU hours for CIFAR datasets. Inspired by \cite{mattern2023membership}, we propose A-LiRA, an augmentation-based LiRA algorithm. Instead of training $2n$ shadow models, A-LiRA only trains \emph{one} in-training shadow model $\mathcal{M}(D|x \in D)$ and one out-training shadow model $\mathcal{M}(D \setminus x)$, applying $n$ augmented samples to observe the output distributions.

A-LiRA is detailed in Algorithm~\ref{alg:A-LiRA} and operates in the following three phases.

\begin{algorithm}[h]
  \caption{Augmentation-based LiRA}\label{alg:A-LiRA}
  \begin{algorithmic}[1]    
    \Require target model $f_T$, example $(x,y)$, dataset $D$, number of augmentations $n$
    \State $\text{obs}_{\text{in}} \gets \{\}$
    \State $\text{obs}_{\text{out}} \gets \{\}$
    \State $\text{obs}_{\text{target}} \gets \{\}$
    \State $f_{\text{in}} \gets \mathcal{M}(D)$ \Comment{Train shadow IN model}
    \State $f_{\text{out}} \gets \mathcal{M}(D \setminus \{(x,y)\})$ \Comment{Train shadow OUT model}
    \State $X_{\text{aug}} \gets \text{Augmentation}(x,n)$ \Comment{Augment the data}
    \For{$x_i \in X_{\text{aug}}$}
      \State $\text{obs}_{\text{in}} \gets \text{obs}_{\text{in}} \cup \{\phi(f_{\text{in}}(x_i)_y)\}$
      \State $\text{obs}_{\text{out}} \gets \text{obs}_{\text{out}} \cup \{\phi(f_{\text{out}}(x_i)_y)\}$
      \State $\text{obs}_{\text{target}} \gets \text{obs}_{\text{target}} \cup \{\phi(f_T(x_i)_y)\}$
    \EndFor
    \State $\mu_{\text{in}} \gets \operatorname{mean}(\text{obs}_{\text{in}})$
    \State $\mu_{\text{out}} \gets \operatorname{mean}(\text{obs}_{\text{out}})$
    \State $\sigma_{\text{in}}^2 \gets \operatorname{var}(\text{obs}_{\text{in}})$
    \State $\sigma_{\text{out}}^2 \gets \operatorname{var}(\text{obs}_{\text{out}})$
    \State Determine $\tau$ through thresholding.
    \State $\Lambda \gets 
      \dfrac{
        PDF\bigl(\max(\text{obs}_{\text{target}})\mid \mathcal{N}(\mu_{\text{in}},\sigma_{\text{in}}^2)\bigr)
      }{
        PDF\bigl(\max(\text{obs}_{\text{target}})\mid \mathcal{N}(\mu_{\text{out}},\sigma_{\text{out}}^2)\bigr)
      }$
    \State \Return “member” if $\Lambda > \tau$ else “non-member”
  \end{algorithmic}
\end{algorithm}

\textbf{Phase 1: Distribution estimation.} For a target sample $(x, y)$, we generate $n$ augmentations $X_{\text{aug}}$ by randomly flipping, rotating, and shifting $x$. We then train two shadow models: an \textit{in}-model $f_{\text{in}} = \mathcal{M}(D)$ (trained with $x$ in the dataset) and an \textit{out}-model $f_{\text{out}} = \mathcal{M}(D \setminus x)$.

For each augmented sample $x_i \in X_{\text{aug}}$, we input it into each of the two models $f_{\text{in}}$ and $f_{\text{out}}$ and obtain the confidence vector $l_i$, which contains $c$ elements for a $c$-class prediction, with $\sum_c l_{i,c} = 1$. We focus on the confidence of the true label $y$, denoted as $l_{i,y}$ for distribution estimation. Following \cite{carlini2022Lira}, we apply a logit transformation to $l_{i,y}$:
\[
\phi(p) = \log\left(\frac{p}{1 - p}\right), \quad \text{where } p = l_{i,y}.
\]
The transformed confidence values are assumed to follow two normal distributions $\mathcal{N}(\mu_{\text{in or out}}, \sigma_{\text{in or out}}^2)$, with parameters estimated from the collection of $\{\phi(l_{i,y})\}$. Typically, $f_{\text{in}}$ will have a higher mean and smaller variance. The output of this phase is the two normal distributions: $\mathcal{N}(\mu_{\text{in}}, \sigma_{\text{in}}^2)$ and $N(\mu_{\text{out}}, \sigma_{\text{out}}^2)$.

\begin{table*}[h]
\centering
\begin{tabular}{|l|llllll|} 
\hline
\multirow{2}{*}{Method} & \multicolumn{3}{l}{CIFAR-10}                     & \multicolumn{3}{l|}{CIFAR-100}                    \\ 
\cline{2-7}
                        & AUC            & TPR@1\%FPR      & Time(hours)          & AUC            & TPR@1\%FPR      & Time(hours)           \\ 
\hline
Online-LiRA             & \textbf{0.706} & 9.7\%           & 4.76          & 0.913          & 25.4\%          & 4.84           \\
Offline-LiRA            & 0.663          & 8.6\%           & \textbf{0.17} & 0.833          & 19.3\%          & \textbf{0.13}  \\ 
\hline
A-LiRA(Ours)            & 0.703          & \textbf{10.3\%} & 0.53          & \textbf{0.917} & \textbf{26.8\%} & 0.57           \\
\hline
\end{tabular}
\caption{Attacking performance on CIFAR datasets. TPR@1\%FPR indicates TPR when FPR is 0.01. Time is the GPU hours to generate classification for one sample, including training shadow models and thresholding. The bolded cell is the best performance.}
\label{tab:attacking}
\end{table*}
\textbf{Phase 2: Making decision.} To determine whether $x$ was used to train the target model $f_T = \mathcal{M}(D)$, the augmented set $X_{\text{aug}}$ is passed through the model, and the same logit transformation is applied to the output confidence values. Intuitively, if $f_T$ was trained on $x$, the transformed confidence values $\phi(l_{i, y})$ should vary less compared to when $f_T$ was not trained on $x$. The maximum value of $\phi(l_{i, y})$ highlights the largest difference between these two cases. We then compute the likelihood of $\max\{\phi(l_{i, y})\}$ follows either of the two normal distributions $\mathcal{N}(\mu_{\text{in}}, \sigma_{\text{in}}^2)$ or $\mathcal{N}(\mu_{\text{out}}, \sigma_{\text{out}}^2)$. As shown by \cite{carlini2022Lira}, the best true positive rate (TPR) at a given false positive rate (FPR) can be found by thresholding the likelihood ratio $\Lambda$:
\[
\Lambda = \frac{PDF(\max\{\phi(l_{i, y})\} | \mathcal{N}(\mu_{\text{in}}, \sigma_{\text{in}}^2))}{PDF(\max\{\phi(l_{i, y})\} | \mathcal{N}(\mu_{\text{out}}, \sigma_{\text{out}}^2))}.
\]
where PDF is the probability density function. This phase outputs the computed $\Lambda$ for the target model $f_T$ and sample $x$. Assuming we have an ideal threshold $\tau$, if $\Lambda > \tau$, we conclude that $f_T$ was trained on $x$; otherwise, it was not.

\textbf{Phase 3: Generating threshold $\tau$ at a specific FPR. (Line 16 in Algorithm~\ref{alg:A-LiRA})} In our paper, we compare A-LiRA and other MIAs on dataset-level (Table~\ref{tab:attacking}). To evaluate the TPR at a specific FPR, we simply set the $\tau$ as a $\Lambda$ of a specific non-member (negative) sample. For instance, if $\text{FPR}=0.01$, then we descendingly sort the $\Lambda$ of non-member samples and choose the top-0.01 $\Lambda$ as the threshold, which makes 0.01 negative samples predicted as positive. 

Except Table~\ref{tab:attacking}, all our experiments estimate the per-sample $\text{TPR}/\text{FPR}$. Specifically, we train $m$ models on random halves of $D$, which means each sample is used to train $m/2$ models. For each sample, we are able to have $m/2$ non-member and $m/2$ member $\Lambda$. Then we use the same thresholding strategy to compute the greatest $\text{TPR}/\text{FPR}$ for each sample.

\section{Experiments}
\label{sec:Exp}
This section presents extensive experiments aimed at addressing the following key questions: 1. How does A-LiRA perform in terms of both efficacy and efficiency compared to the LiRA methods? 2.  How well do the published unlearning methods meet the proposed auditing measures?

\subsection{Setup}

Before exploring these questions, the experimental setup is outlined.

\textbf{Datasets and models.} We use CIFAR-10 and CIFAR-100 to train ResNet-18 models with FFCV (\cite{leclerc2023ffcv}), using a learning rate of 0.5, weight decay of 5e-4, and training for 50 epochs with early stopping on NVIDIA Titan V-100 and Titan Xp GPUs. Each sample has 10 random augmentations to enhance generalization. We train differentially-private models (DP models) using DP-SGD\cite{abadi2016deep}. Due to computational constraints, we limit our experiments to these datasets, as large-scale datasets would require training multiple models for privacy risk estimation, which is impractical with current resources.

\textbf{Approximate unlearning methods.} We choose SUNSHINE \cite{golatkar2020eternal}, SSD \cite{foster2024fast}, and SalUn \cite{fan2024salun} as the representative approximate unlearning methods as they are the latest algorithms with the best performance reported.

\textbf{Membership inference attacks.} We use A-LiRA as the attack method, generating 100 augmentations per sample via random flipping, rotation, shifting, etc. One shadow \textit{in}-model and one shadow \textit{out}-model are trained to estimate the distributions. We then train 30 \textit{in}- and \textit{out}-models to determine the threshold. All models use the same hyperparameters and data distribution as the target model. We also show the results of privacy sensitivity estimation using other attacks in Appendix B.

\textbf{Estimating per-sample privacy risk.} We train 500 models on random halves of the entire training set to ensure each $x \in D$ is used to train around 250 models. We then compute per-sample $\text{TPR}_x/\text{FPR}_x$ on these models. To estimate the privacy risks of unlearned and retained samples after unlearning, we simply unlearn the unlearning samples from the models that were trained on unlearning samples, as detailed in Section~\ref{sec:prelimineries}.

\textbf{Metrics.} We use the Area Under the ROC Curve (AUC) score, $\text{TPR}$ when $\text{FPR}=0.01$, and GPU time to evaluate membership inference attacks. We use the ratio of samples that failed to meet Criteria 1\&2 to evaluate the privacy protection of machine unlearning methods. Specifically, we check the ratio of unlearned samples that do not meet Criterion 1 and the ratio of retained samples that do not meet Criterion 2, denoted as the ``Failure Rate’’ in the figures.

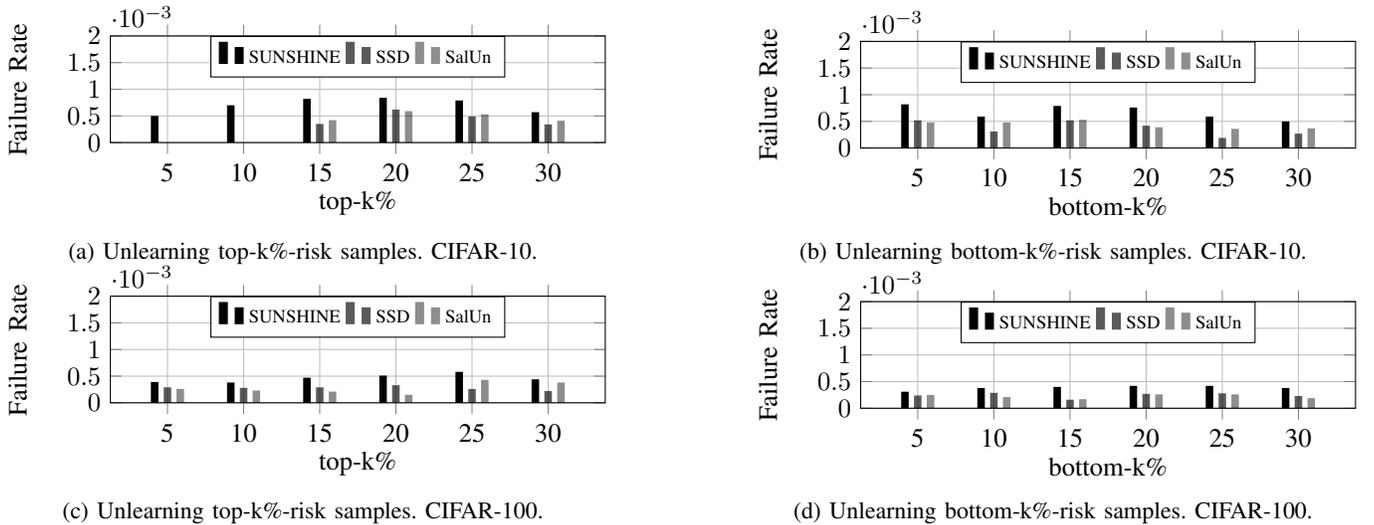
\begin{figure*}[h]
    \centering
    \begin{subfigure}[b]{0.45\textwidth}
        \centering
        \begin{tikzpicture}
            \begin{axis}[
                width=\textwidth, height=3cm,
                xlabel={top-k\%},
                ylabel={Failure Rate},
                ymax=0.002,
                ymin=0,
                grid=both,
               legend style={at={(0.5,1)}, anchor=north,legend columns=-1,font=\scriptsize},
                ybar, 
                xtick=data, 
                bar width=0.1cm, 
                enlarge x limits=0.15, 
                xticklabels={5,10,15,20,25,30}, 
                symbolic x coords={0.05,0.1,0.15,0.2,0.25,0.3}, 
                xticklabels={5,10,15,20,25,30}
            ]

	\addplot[color=black, fill=black, draw=none,opacity=1] table[x=k, y expr=1-\thisrow{SUNSHINE}, col sep=comma] {q1_approximate_cifar10_largest.csv};
	\addlegendentry{SUNSHINE}
	\addplot[color=black, fill=black, draw=none,opacity=0.65] table[x=k, y expr=1-\thisrow{SSD}, col sep=comma] {q1_approximate_cifar10_largest.csv};
	\addlegendentry{SSD}
	\addplot[color=black,fill=black,draw=none,opacity=0.45] table[x=k, y expr=1-\thisrow{SalUn}, col sep=comma] {q1_approximate_cifar10_largest.csv};
	\addlegendentry{SalUn}
	\end{axis}
        \end{tikzpicture}
        \caption{Unlearning top-k\%-risk samples. CIFAR-10.}
        \label{fig:top-down-approximate-q1-10}
    \end{subfigure}%
    \hfill
        \begin{subfigure}[b]{0.45\textwidth}
        \centering
        \begin{tikzpicture}
            \begin{axis}[
                width=\textwidth, height=3cm,
                xlabel={bottom-k\%},
                ylabel={Failure Rate},
                ymax=0.002,
                ymin=0,
                grid=both,
               legend style={at={(0.5,1)}, anchor=north,legend columns=-1,font=\scriptsize},
                ybar, 
                xtick=data, 
                bar width=0.1cm, 
                enlarge x limits=0.15, 
                xticklabels={5,10,15,20,25,30}, 
                symbolic x coords={0.05,0.1,0.15,0.2,0.25,0.3}, 
                xticklabels={5,10,15,20,25,30}
            ]

	\addplot[color=black, fill=black, draw=none,opacity=1] table[x=k, y expr=1-\thisrow{SUNSHINE}, col sep=comma] {q1_approximate_cifar10_smallest.csv};
	\addlegendentry{SUNSHINE}
	\addplot[color=black, fill=black, draw=none,opacity=0.65] table[x=k, y expr=1-\thisrow{SSD}, col sep=comma] {q1_approximate_cifar10_smallest.csv};
	\addlegendentry{SSD}
	\addplot[color=black,fill=black,draw=none,opacity=0.45] table[x=k, y expr=1-\thisrow{SalUn}, col sep=comma] {q1_approximate_cifar10_smallest.csv};
	\addlegendentry{SalUn}
	\end{axis}
        \end{tikzpicture}
        \caption{Unlearning bottom-k\%-risk samples. CIFAR-10.}
        \label{fig:bottom-up-approximate-q1-10}
    \end{subfigure}%
    \hfill
        \begin{subfigure}[b]{0.45\textwidth}
        \centering
        \begin{tikzpicture}
            \begin{axis}[
                width=\textwidth, height=3cm,
                xlabel={top-k\%},
                ylabel={Failure Rate},
                ymax=0.002,
                ymin=0,
                grid=both,
               legend style={at={(0.5,1)}, anchor=north,legend columns=-1,font=\scriptsize},
                ybar, 
                xtick=data, 
                bar width=0.1cm, 
                enlarge x limits=0.15, 
                xticklabels={5,10,15,20,25,30}, 
                symbolic x coords={0.05,0.1,0.15,0.2,0.25,0.3}, 
                xticklabels={5,10,15,20,25,30}
            ]

            \addplot[color=black, fill=black,draw=none, opacity=1] table[x=k, y expr=1-\thisrow{SUNSHINE}, col sep=comma] {q1_approximate_cifar100_largest.csv};
            \addlegendentry{SUNSHINE}
            \addplot[color=black, fill=black,draw=none, opacity=0.65] table[x=k, y expr=1-\thisrow{SSD}, col sep=comma] {q1_approximate_cifar100_largest.csv};
            \addlegendentry{SSD}
            \addplot[color=black,fill=black,draw=none,opacity=0.45] table[x=k, y expr=1-\thisrow{SalUn}, col sep=comma] {q1_approximate_cifar100_largest.csv};
            \addlegendentry{SalUn}
            \end{axis}
        \end{tikzpicture}
        \caption{Unlearning top-k\%-risk samples. CIFAR-100.}
        \label{fig:top-down-approximate-q1-100}
    \end{subfigure}
     \hfill
        \begin{subfigure}[b]{0.45\textwidth}
        \centering
        \begin{tikzpicture}
            \begin{axis}[
                width=\textwidth, height=3cm,
                xlabel={bottom-k\%},
                ylabel={Failure Rate},
                ymax=0.002,
                ymin=0,
                grid=both,
               legend style={at={(0.5,1)}, anchor=north,legend columns=-1,font=\scriptsize},
                ybar, 
                xtick=data, 
                bar width=0.1cm, 
                enlarge x limits=0.15, 
                xticklabels={5,10,15,20,25,30}, 
                symbolic x coords={0.05,0.1,0.15,0.2,0.25,0.3}, 
                xticklabels={5,10,15,20,25,30}
            ]

            \addplot[color=black, fill=black,draw=none, opacity=1] table[x=k, y expr=1-\thisrow{SUNSHINE}, col sep=comma] {q1_approximate_cifar100_smallest.csv};
            \addlegendentry{SUNSHINE}
            \addplot[color=black, fill=black,draw=none, opacity=0.65] table[x=k, y expr=1-\thisrow{SSD}, col sep=comma] {q1_approximate_cifar100_smallest.csv};
            \addlegendentry{SSD}
            \addplot[color=black,fill=black,draw=none,opacity=0.45] table[x=k, y expr=1-\thisrow{SalUn}, col sep=comma] {q1_approximate_cifar100_smallest.csv};
            \addlegendentry{SalUn}
            \end{axis}
        \end{tikzpicture}
        \caption{Unlearning bottom-k\%-risk samples. CIFAR-100.}
        \label{fig:bottom-up-approximate-q1-100}
    \end{subfigure}
     \caption{Criterion 1: Approximate unlearning delivers less privacy protection to removed samples, compared to retraining.}
    \label{fig:approximate-q1}
\end{figure*}

\subsection{Attacking Performance}

We present the attacking performance in Table~\ref{tab:attacking}. All models are trained using ResNet-18. For both online and offline-LiRA, we totally train 512 shadow models (in and out), which is suggested by \cite{carlini2022Lira}. In the offline-LiRA setting, we use a batch size of 50, meaning that 50 samples are explicitly used to train the out-models, and the attack is performed on those same samples. While Carlini et al. (\cite{carlini2022Onion}) demonstrated that batch strategies can be applied to both online and offline-LiRA, and that setting the batch size to half the dataset size (i.e., half the data used to train the shadow \textit{in}-model) can be effective, we found this approach reduced attack effectiveness.  

Online-LiRA performed the best in global AUC, while our A-LiRA achieved comparable AUC and better TPR at low FPR with significantly lower time costs. On CIFAR-100, A-LiRA outperformed in both AUC and TPR at low FPR. Although offline-LiRA yielded the lowest performance across all experiments, it substantially reduced time costs, making it a viable alternative when time efficiency is a priority.

\begin{figure*}[h]
    \centering
    \begin{subfigure}[b]{0.45\textwidth}
        \centering
        \begin{tikzpicture}
            \begin{axis}[
                width=\textwidth, height=3cm,
                xlabel={top-k\%},
                ylabel={Failure Rate},
                ymax=0.001,
                ymin=0,
                grid=both,
               legend style={at={(0.5,1)}, anchor=north,legend columns=-1,font=\scriptsize},
                ybar, 
                xtick=data, 
                bar width=0.1cm, 
                enlarge x limits=0.15, 
                xticklabels={5,10,15,20,25,30}, 
                symbolic x coords={0.05,0.1,0.15,0.2,0.25,0.3}, 
                xticklabels={5,10,15,20,25,30}
            ]

	\addplot[color=black, fill=black, draw=none,opacity=1] table[x=k, y expr=1-\thisrow{SUNSHINE}, col sep=comma] {q2_approximate_cifar10_largest_epsilon_2.csv};
	\addlegendentry{SUNSHINE}
	\addplot[color=black, fill=black, draw=none,opacity=0.65] table[x=k, y expr=1-\thisrow{SSD}, col sep=comma] {q2_approximate_cifar10_largest_epsilon_2.csv};
	\addlegendentry{SSD}
	\addplot[color=black,fill=black,draw=none,opacity=0.45] table[x=k, y expr=1-\thisrow{SalUn}, col sep=comma] {q2_approximate_cifar10_largest_epsilon_2.csv};
	\addlegendentry{SalUn}
	\end{axis}
        \end{tikzpicture}
        \caption{Unlearning top-k\%-risk samples. CIFAR-10.}
        \label{fig:top-down-approximate-q1-10-2}
    \end{subfigure}%
    \hfill
        \begin{subfigure}[b]{0.45\textwidth}
        \centering
        \begin{tikzpicture}
            \begin{axis}[
                width=\textwidth, height=3cm,
                xlabel={bottom-k\%},
                ylabel={Failure Rate},
                ymax=0.001,
                ymin=0,
                grid=both,
               legend style={at={(0.5,1)}, anchor=north,legend columns=-1,font=\scriptsize},
                ybar, 
                xtick=data, 
                bar width=0.1cm, 
                enlarge x limits=0.15, 
                xticklabels={5,10,15,20,25,30}, 
                symbolic x coords={0.05,0.1,0.15,0.2,0.25,0.3}, 
                xticklabels={5,10,15,20,25,30}
            ]

	\addplot[color=black, fill=black, draw=none,opacity=1] table[x=k, y expr=1-\thisrow{SUNSHINE}, col sep=comma] {q2_approximate_cifar10_smallest_epsilon_2.csv};
	\addlegendentry{SUNSHINE}
	\addplot[color=black, fill=black, draw=none,opacity=0.65] table[x=k, y expr=1-\thisrow{SSD}, col sep=comma] {q2_approximate_cifar10_smallest_epsilon_2.csv};
	\addlegendentry{SSD}
	\addplot[color=black,fill=black,draw=none,opacity=0.45] table[x=k, y expr=1-\thisrow{SalUn}, col sep=comma] {q2_approximate_cifar10_smallest_epsilon_2.csv};
	\addlegendentry{SalUn}
	\end{axis}
        \end{tikzpicture}
        \caption{Unlearning bottom-k\%-risk samples. CIFAR-10.}
        \label{fig:bottom-up-approximate-q1-10-2}
    \end{subfigure}%
    \hfill
        \begin{subfigure}[b]{0.45\textwidth}
        \centering
        \begin{tikzpicture}
            \begin{axis}[
                width=\textwidth, height=3cm,
                xlabel={top-k\%},
                ylabel={Failure Rate},
                ymax=0.001,
                ymin=0,
                grid=both,
               legend style={at={(0.5,1)}, anchor=north,legend columns=-1,font=\scriptsize},
                ybar, 
                xtick=data, 
                bar width=0.1cm, 
                enlarge x limits=0.15, 
                xticklabels={5,10,15,20,25,30}, 
                symbolic x coords={0.05,0.1,0.15,0.2,0.25,0.3}, 
                xticklabels={5,10,15,20,25,30}
            ]

            \addplot[color=black, fill=black,draw=none, opacity=1] table[x=k, y expr=1-\thisrow{SUNSHINE}, col sep=comma] {q2_approximate_cifar100_largest_epsilon_2.csv};
            \addlegendentry{SUNSHINE}
            \addplot[color=black, fill=black,draw=none, opacity=0.65] table[x=k, y expr=1-\thisrow{SSD}, col sep=comma] {q2_approximate_cifar100_largest_epsilon_2.csv};
            \addlegendentry{SSD}
            \addplot[color=black,fill=black,draw=none,opacity=0.45] table[x=k, y expr=1-\thisrow{SalUn}, col sep=comma] {q2_approximate_cifar100_largest_epsilon_2.csv};
            \addlegendentry{SalUn}
            \end{axis}
        \end{tikzpicture}
        \caption{Unlearning top-k\%-risk samples. CIFAR-100.}
        \label{fig:top-down-approximate-q1-100-2}
    \end{subfigure}
     \hfill
        \begin{subfigure}[b]{0.45\textwidth}
        \centering
        \begin{tikzpicture}
            \begin{axis}[
                width=\textwidth, height=3cm,
                xlabel={bottom-k\%},
                ylabel={Failure Rate},
                ymax=0.001,
                ymin=0,
                grid=both,
               legend style={at={(0.5,1)}, anchor=north,legend columns=-1,font=\scriptsize},
                ybar, 
                xtick=data, 
                bar width=0.1cm, 
                enlarge x limits=0.15, 
                xticklabels={5,10,15,20,25,30}, 
                symbolic x coords={0.05,0.1,0.15,0.2,0.25,0.3}, 
                xticklabels={5,10,15,20,25,30}
            ]

            \addplot[color=black, fill=black,draw=none, opacity=1] table[x=k, y expr=1-\thisrow{SUNSHINE}, col sep=comma] {q2_approximate_cifar100_smallest_epsilon_2.csv};
            \addlegendentry{SUNSHINE}
            \addplot[color=black, fill=black,draw=none, opacity=0.65] table[x=k, y expr=1-\thisrow{SSD}, col sep=comma] {q2_approximate_cifar100_smallest_epsilon_2.csv};
            \addlegendentry{SSD}
            \addplot[color=black,fill=black,draw=none,opacity=0.45] table[x=k, y expr=1-\thisrow{SalUn}, col sep=comma] {q2_approximate_cifar100_smallest_epsilon_2.csv};
            \addlegendentry{SalUn}
            \end{axis}
        \end{tikzpicture}
        \caption{Unlearning bottom-k\%-risk samples. CIFAR-100.}
        \label{fig:bottom-up-approximate-q1-100-2}
    \end{subfigure}
     \caption{Criterion 2: Approximate unlearning increases the privacy risk of some retained samples in differnetially private models where $\epsilon=2$.}
    \label{fig:approximate-ep2}
\end{figure*}

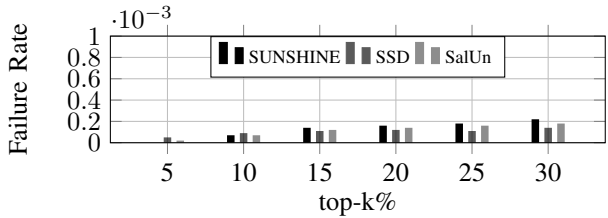
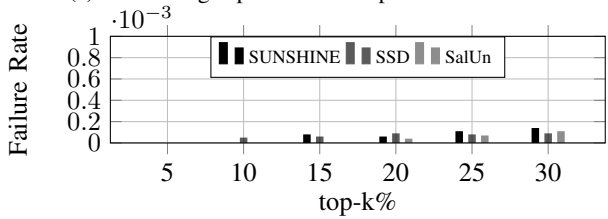
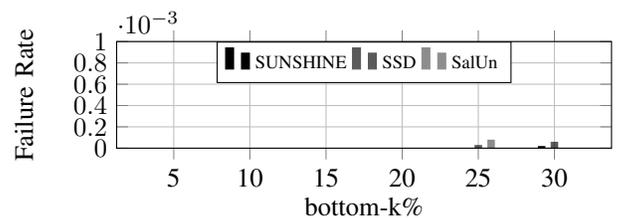
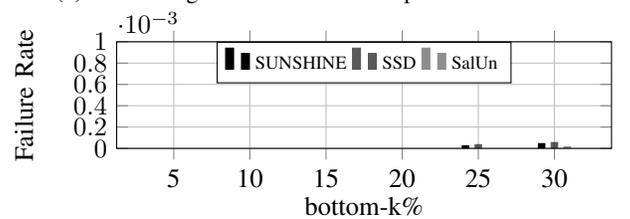
\begin{figure*}[h]
    \centering
    \begin{subfigure}[b]{0.45\textwidth}
        \centering
        \begin{tikzpicture}
            \begin{axis}[
                width=\textwidth, height=3cm,
                xlabel={top-k\%},
                ylabel={Failure Rate},
                ymax=0.001,
                ymin=0,
                grid=both,
               legend style={at={(0.5,1)}, anchor=north,legend columns=-1,font=\scriptsize},
                ybar, 
                xtick=data, 
                bar width=0.1cm, 
                enlarge x limits=0.15, 
                xticklabels={5,10,15,20,25,30}, 
                symbolic x coords={0.05,0.1,0.15,0.2,0.25,0.3}, 
                xticklabels={5,10,15,20,25,30}
            ]

	\addplot[color=black, fill=black, draw=none,opacity=1] table[x=k, y expr=1-\thisrow{SUNSHINE}, col sep=comma] {q2_approximate_cifar10_largest_epsilon_6.csv};
	\addlegendentry{SUNSHINE}
	\addplot[color=black, fill=black, draw=none,opacity=0.65] table[x=k, y expr=1-\thisrow{SSD}, col sep=comma] {q2_approximate_cifar10_largest_epsilon_6.csv};
	\addlegendentry{SSD}
	\addplot[color=black,fill=black,draw=none,opacity=0.45] table[x=k, y expr=1-\thisrow{SalUn}, col sep=comma] {q2_approximate_cifar10_largest_epsilon_6.csv};
	\addlegendentry{SalUn}
	\end{axis}
        \end{tikzpicture}
        \caption{Unlearning top-k\%-risk samples. CIFAR-10.}
        \label{fig:top-down-approximate-q1-10-6}
    \end{subfigure}%
    \hfill
        \begin{subfigure}[b]{0.45\textwidth}
        \centering
        \begin{tikzpicture}
            \begin{axis}[
                width=\textwidth, height=3cm,
                xlabel={bottom-k\%},
                ylabel={Failure Rate},
                ymax=0.001,
                ymin=0,
                grid=both,
               legend style={at={(0.5,1)}, anchor=north,legend columns=-1,font=\scriptsize},
                ybar, 
                xtick=data, 
                bar width=0.1cm, 
                enlarge x limits=0.15, 
                xticklabels={5,10,15,20,25,30}, 
                symbolic x coords={0.05,0.1,0.15,0.2,0.25,0.3}, 
                xticklabels={5,10,15,20,25,30}
            ]

	\addplot[color=black, fill=black, draw=none,opacity=1] table[x=k, y expr=1-\thisrow{SUNSHINE}, col sep=comma] {q2_approximate_cifar10_smallest_epsilon_6.csv};
	\addlegendentry{SUNSHINE}
	\addplot[color=black, fill=black, draw=none,opacity=0.65] table[x=k, y expr=1-\thisrow{SSD}, col sep=comma] {q2_approximate_cifar10_smallest_epsilon_6.csv};
	\addlegendentry{SSD}
	\addplot[color=black,fill=black,draw=none,opacity=0.45] table[x=k, y expr=1-\thisrow{SalUn}, col sep=comma] {q2_approximate_cifar10_smallest_epsilon_6.csv};
	\addlegendentry{SalUn}
	\end{axis}
        \end{tikzpicture}
        \caption{Unlearning bottom-k\%-risk samples. CIFAR-10.}
        \label{fig:bottom-up-approximate-q1-10-6}
    \end{subfigure}%
    \hfill
        \begin{subfigure}[b]{0.45\textwidth}
        \centering
        \begin{tikzpicture}
            \begin{axis}[
                width=\textwidth, height=3cm,
                xlabel={top-k\%},
                ylabel={Failure Rate},
                ymax=0.001,
                ymin=0,
                grid=both,
               legend style={at={(0.5,1)}, anchor=north,legend columns=-1,font=\scriptsize},
                ybar, 
                xtick=data, 
                bar width=0.1cm, 
                enlarge x limits=0.15, 
                xticklabels={5,10,15,20,25,30}, 
                symbolic x coords={0.05,0.1,0.15,0.2,0.25,0.3}, 
                xticklabels={5,10,15,20,25,30}
            ]

            \addplot[color=black, fill=black,draw=none, opacity=1] table[x=k, y expr=1-\thisrow{SUNSHINE}, col sep=comma] {q2_approximate_cifar100_largest_epsilon_6.csv};
            \addlegendentry{SUNSHINE}
            \addplot[color=black, fill=black,draw=none, opacity=0.65] table[x=k, y expr=1-\thisrow{SSD}, col sep=comma] {q2_approximate_cifar100_largest_epsilon_6.csv};
            \addlegendentry{SSD}
            \addplot[color=black,fill=black,draw=none,opacity=0.45] table[x=k, y expr=1-\thisrow{SalUn}, col sep=comma] {q2_approximate_cifar100_largest_epsilon_6.csv};
            \addlegendentry{SalUn}
            \end{axis}
        \end{tikzpicture}
        \caption{Unlearning top-k\%-risk samples. CIFAR-100.}
        \label{fig:top-down-approximate-q1-100-6}
    \end{subfigure}
     \hfill
        \begin{subfigure}[b]{0.45\textwidth}
        \centering
        \begin{tikzpicture}
            \begin{axis}[
                width=\textwidth, height=3cm,
                xlabel={bottom-k\%},
                ylabel={Failure Rate},
                ymax=0.001,
                ymin=0,
                grid=both,
               legend style={at={(0.5,1)}, anchor=north,legend columns=-1,font=\scriptsize},
                ybar, 
                xtick=data, 
                bar width=0.1cm, 
                enlarge x limits=0.15, 
                xticklabels={5,10,15,20,25,30}, 
                symbolic x coords={0.05,0.1,0.15,0.2,0.25,0.3}, 
                xticklabels={5,10,15,20,25,30}
            ]

            \addplot[color=black, fill=black,draw=none, opacity=1] table[x=k, y expr=1-\thisrow{SUNSHINE}, col sep=comma] {q2_approximate_cifar100_smallest_epsilon_6.csv};
            \addlegendentry{SUNSHINE}
            \addplot[color=black, fill=black,draw=none, opacity=0.65] table[x=k, y expr=1-\thisrow{SSD}, col sep=comma] {q2_approximate_cifar100_smallest_epsilon_6.csv};
            \addlegendentry{SSD}
            \addplot[color=black,fill=black,draw=none,opacity=0.45] table[x=k, y expr=1-\thisrow{SalUn}, col sep=comma] {q2_approximate_cifar100_smallest_epsilon_6.csv};
            \addlegendentry{SalUn}
            \end{axis}
        \end{tikzpicture}
        \caption{Unlearning bottom-k\%-risk samples. CIFAR-100.}
        \label{fig:bottom-up-approximate-q1-100-6}
    \end{subfigure}
     \caption{Criterion 2: Approximate unlearning increases the privacy risk of some retained samples in differentially private models where $\epsilon=4$.}
    \label{fig:approximate-ep6}
\end{figure*}

\subsection{Does unlearning protect privacy?}
In this section, we investigate whether existing approximate machine unlearning satisfactorily protects privacy according to the proposed criteria. 

To investigate how these approximate unlearning methods perform on samples with different levels of privacy risk. We first estimate the privacy risk of each sample in $D$ on model $M_D$ with A-LiRA. Then, we try two batches of removal: top-$k\%$, the most risky samples, and bottom-$k\%$, the least risky ones. For each type of removal, we apply approximate unlearning algorithms and evaluate the samples' privacy risk under the new model.

\textbf{Criterion 1.} A straightforward question is whether the unlearning strategies can protect the privacy of unlearned samples. To evaluate this question, we evaluate the failure rate of Criterion 1 on both non-DP models and DP models. Figure~\ref{fig:approximate-q1} shows the results on non-DP models. While most unlearned samples become more private, a small part of the unlearned samples become more sensitive even after removal. These counterintuitive results indicate that unlearning may not always protect the privacy of unlearned samples. According to Carlini et al. \cite{carlini2022Onion}, the existence of duplicated samples will mitigate the sensitivity of each of the duplicates. We hypothesize that the approximate unlearning methods may unlearn part of the memorized features, which do the deduplication for the model and increase the sensitivity of some left features. We also observed that unlearning to protect privacy is more effective on some sensitive samples (Figure~\ref{fig:top-down-approximate-q1-10} and \ref{fig:top-down-approximate-q1-100}) while less effective on some originally non-sensitive samples (Figure~\ref{fig:bottom-up-approximate-q1-10} and \ref{fig:bottom-up-approximate-q1-100}). This observation also supports the hypothesis. Some originally non-sensitive inliers shared similar features, and approximate unlearning may remove part of them. However, this hypothesis is not the main topic of this paper, we will leave this to our future work. 

In contrast, in the DP model, when $\epsilon$ is 2 and 6, we observed that all unlearned samples become safer with decreased $\text{TPR}/\text{FPR}$ after unlearning (we don't show the results as a blank figure). Thus, we think approximate machine unlearning may be more effective to protect the privacy of unlearned samples when the model is differentially private.

\textbf{Criterion 2.} The results of Criterion 1 in differentially private models show that approximate unlearning can effectively decrease the privacy risks of unlearned samples. However, in a DP model, estimated $\text{TPR}/\text{FPR}>\epsilon$ for even one sample results in the breaking of differentially private and user-agreements of privacy. Thus, we are curious about how approximate machine unlearning affects the privacy of retained samples. We train the models with $\epsilon=2,6$ using DP-SGD \cite{abadi2016deep}. Similarly, we want to observe the impact of unlearning top-k\% or bottom-k\%-risk samples, respectively, and check how many retained samples dissatisfy Criterion 2 (Equation~\ref{eq:2}). Figure~\ref{fig:approximate-ep2} shows that all three unlearning methods will increase the privacy risk of some retained samples when $\epsilon =2$. Unlearning modes sensitive samples (Figure~\ref{fig:top-down-approximate-q1-10-2} and \ref{fig:top-down-approximate-q1-100-2}) results in more samples breaking the differential privacy compared to unlearning non-sensitive samples (Figure~\ref{fig:bottom-up-approximate-q1-10-2} and \ref{fig:bottom-up-approximate-q1-100-2}). This pattern also exists when we use a more relaxed differential privacy, i.e., $\epsilon=6$, as shown in Figure~\ref{fig:approximate-ep6}. This pattern is consistent with the ``privacy onion effect’’ (\cite{carlini2022Onion}), which indicates that removing some sensitive samples will expose some non-sensitive samples. We also observe that this phenomenon may be reduced with a more relaxed setting of $\epsilon$. This makes sense because of the higher upper bound of $\text{TPR}/\text{FPR}$. Note that for a non-DP model, although we have observed some retained samples become more sensitive, this may not break user agreements because there is no strict upper bound for privacy risk. For exploration, we design an alternative upper bound for non-DP models in Appendix A and show that the retained samples will still exceed the upper bound.

\textbf{Discussion.} Using A-LiRA as the estimator, we have found several interesting results. Firstly, existing approximate machine unlearning may not provide sufficient privacy protection for unlearned samples. We hypothesize it because of the duplication of unlearning samples in the training set. The approximate unlearning may unlearn some features and thus make the remaining features become outliers. Secondly, machine unlearning may also increase the privacy risk of retained samples, which can be critical if differentially private machine learning is applied. Thus, it's vital to reconsider how to design proper machine unlearning methods to comprehensively protect the privacy of all involved data samples. Our A-LiRA estimator and acceptance matrix can be a powerful tool to help evaluate such machine unlearning methods.

\section{Conclusion}

The privacy protection criteria for machine unlearning have not been sufficiently studied yet. We propose two criteria to audit the privacy risks of unlearned and retained samples to fully understand a machine unlearning algorithm’s privacy protection capacity. The core of the proposed auditing mechanism is an efficient sample-level membership inference attack, A-LiRA. We show in experiments that A-LiRA performs efficiently with comparable attacking accuracy to the original online-LiRA algorithm, making it deployable to real applications. With the proposed criteria, we also show that most existing machine unlearning algorithms do not satisfactorily protect samples’ privacy. 

\clearpage
\bibliography{ref}
\bibliographystyle{acm}

\clearpage

\section*{Appendix A: Criterion 2 of Non-differentially Private Models}
\label{ap:a}
In this section, we introduce an alternative to Equation~\ref{eq:2} in non-DP private models. Specifically, we design the upper bound of per-sample sensitivity after unlearning as the maximum sensitivity of the original model:
\begin{equation*}
   E(U_{{M_D},X}, x|x\in D \setminus X) \leq \max(E(M_D, x | x\in D)) + t_2
\end{equation*}
The intuition behind this upper bound is that the retained samples' sensitivity should not become more sensitive than the most sensitive sample of the original model. $t_2$ is a relaxation for this upper bound that can be adjusted by the model builder according to the user agreements. When $t_2>0$, the upper bound becomes loose and tight otherwise. This upper bound only shows an example of how to measure Criterion 2 in non-DP models. The model builders can design their upper bounds accordingly.

\begin{figure}[h]
    \centering
    \begin{subfigure}[b]{0.48\textwidth}
        \centering
        \begin{tikzpicture}
            \begin{axis}[
                width=\textwidth, height=3cm,
                xlabel={top-k\%},
                ylabel={Failure Rate},
                ymax=0.06,
                ymin=0,
                grid=both,
               legend style={at={(0.5,1)}, anchor=north,legend columns=-1,font=\scriptsize},
                ybar, 
                xtick=data, 
                bar width=0.1cm, 
                enlarge x limits=0.15, 
                xticklabels={5,10,15,20,25,30}, 
                symbolic x coords={0.05,0.1,0.15,0.2,0.25,0.3}, 
                xticklabels={5,10,15,20,25,30}
            ]

	\addplot[color=black, fill=black, draw=none,opacity=1] table[x=k, y expr=1-\thisrow{SUNSHINE}, col sep=comma] {q2_approximate_cifar100_largest.csv};
	\addlegendentry{SUNSHINE}

	\addplot[color=black, fill=black, draw=none,opacity=0.65] table[x=k, y expr=1-\thisrow{SSD}, col sep=comma] {q2_approximate_cifar100_largest.csv};
	\addlegendentry{SSD}

	\addplot[color=black,fill=black,draw=none,opacity=0.45] table[x=k, y expr=1-\thisrow{SalUn}, col sep=comma] {q2_approximate_cifar100_largest.csv};
	\addlegendentry{SalUn}
	
	\end{axis}

        \end{tikzpicture}
        \caption{Unlearning top-k\%-risk samples. CIFAR-10.}
        \label{fig:top-down-approximate-q2-10-ap}
    \end{subfigure}%
    \hfill
    \begin{subfigure}[b]{0.48\textwidth}
        \centering
        \begin{tikzpicture}
            \begin{axis}[
                width=\textwidth, height=3cm,
                xlabel={bottom-k\%},
                ylabel={Failure Rate},
                ymax=0.06,
                ymin=0,
                grid=both,
               legend style={at={(0.5,1)}, anchor=north,legend columns=-1,font=\scriptsize},
                ybar, 
                xtick=data, 
                bar width=0.1cm, 
                enlarge x limits=0.15, 
                xticklabels={5,10,15,20,25,30}, 
                symbolic x coords={0.05,0.1,0.15,0.2,0.25,0.3}, 
                xticklabels={5,10,15,20,25,30}
            ]

            \addplot[color=black, fill=black,draw=none,opacity=1] table[x=k, y expr=1-\thisrow{SUNSHINE}, col sep=comma] {q2_approximate_cifar10_smallest.csv};
            \addlegendentry{SUNSHINE}
             \addplot[color=black, fill=black,draw=none, opacity=0.65] table[x=k, y expr=1-\thisrow{SSD}, col sep=comma] {q2_approximate_cifar10_smallest.csv};
            \addlegendentry{SSD}
            \addplot[color=black,fill=black,draw=none,opacity=0.45] table[x=k, y expr=1-\thisrow{SalUn}, col sep=comma] {q2_approximate_cifar10_smallest.csv};
            \addlegendentry{SalUn}
            \end{axis}
        \end{tikzpicture}
        \caption{Unlearning bottom-k\%-risk samples. CIFAR-10.}
        \label{fig:bottom-up-approximate-q2-10-ap}
    \end{subfigure}
    \hfill
        \begin{subfigure}[b]{0.48\textwidth}
        \centering
        \begin{tikzpicture}
            \begin{axis}[
                width=\textwidth, height=3cm,
                xlabel={top-k\%},
                ylabel={Failure Rate},
                ymax=0.06,
                ymin=0,
                grid=both,
               legend style={at={(0.5,1)}, anchor=north,legend columns=-1,font=\scriptsize},
                ybar, 
                xtick=data, 
                bar width=0.1cm, 
                enlarge x limits=0.15, 
                xticklabels={5,10,15,20,25,30}, 
                symbolic x coords={0.05,0.1,0.15,0.2,0.25,0.3}, 
                xticklabels={5,10,15,20,25,30}
            ]

            \addplot[color=black, fill=black,draw=none, opacity=1] table[x=k, y expr=1-\thisrow{SUNSHINE}, col sep=comma] {q2_approximate_cifar100_largest.csv};
            \addlegendentry{SUNSHINE}
             \addplot[color=black, fill=black,draw=none, opacity=0.65] table[x=k, y expr=1-\thisrow{SSD}, col sep=comma] {q2_approximate_cifar100_largest.csv};
            \addlegendentry{SSD}
            \addplot[color=black,fill=black,draw=none,opacity=0.45] table[x=k, y expr=1-\thisrow{SalUn}, col sep=comma] {q2_approximate_cifar100_largest.csv};
            \addlegendentry{SalUn}
            \end{axis}
        \end{tikzpicture}
        \caption{Unlearning top-k\%-risk samples. CIFAR-100.}
        \label{fig:top-down-approximate-q2-100-ap}
    \end{subfigure}
    \hfill
        \begin{subfigure}[b]{0.48\textwidth}
        \centering
        \begin{tikzpicture}
            \begin{axis}[
                width=\textwidth, height=3cm,
                xlabel={bottom-k\%},
                ylabel={Failure Rate},
                ymax=0.06,
                ymin=0,
                grid=both,
               legend style={at={(0.5,1)}, anchor=north,legend columns=-1,font=\scriptsize},
                ybar, 
                xtick=data, 
                bar width=0.1cm, 
                enlarge x limits=0.15, 
                xticklabels={5,10,15,20,25,30}, 
                symbolic x coords={0.05,0.1,0.15,0.2,0.25,0.3}, 
                xticklabels={5,10,15,20,25,30}
            ]
            \addplot[color=black, fill=black, draw=none,opacity=1] table[x=k, y expr=1-\thisrow{SUNSHINE}, col sep=comma] {q2_approximate_cifar100_smallest.csv};
            \addlegendentry{SUNSHINE}
             \addplot[color=black, fill=black, draw=none,opacity=0.65] table[x=k, y expr=1-\thisrow{SSD}, col sep=comma] {q2_approximate_cifar100_smallest.csv};
            \addlegendentry{SSD}
            \addplot[color=black,fill=black,draw=none,opacity=0.45] table[x=k, y expr=1-\thisrow{SalUn}, col sep=comma] {q2_approximate_cifar100_smallest.csv};
            \addlegendentry{SalUn}
            \end{axis}
        \end{tikzpicture}
        \caption{Unlearning bottom-k\%-risk samples.CIFAR-100.}
        \label{fig:bottom-up-approximate-q2-100-ap}
    \end{subfigure}
    \caption{Criterion 2: Approximate unlearning breaches the privacy of retained samples in non-DP models. }
    \label{fig:approximate-q2-ap}
\end{figure}
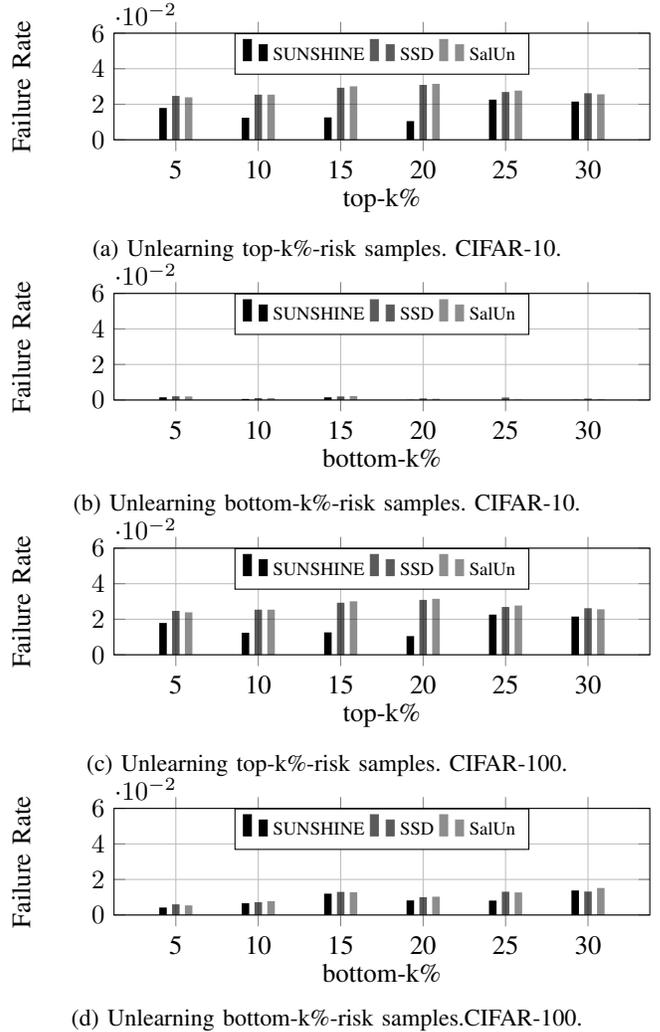

We set $t_2=0$ and see how non-DP models satisfy Criterion 2. As shown in Figure~\ref{fig:approximate-q2-ap}, non-DP models show that a significant amount of retained samples become more sensitive. Unlearning the most sensitive samples will expose more originally non-sensitive samples compared to unlearning the least sensitive ones. This observation aligns with which in DP-models and follows the "privacy onion effect" \cite{carlini2022Onion}.

\section*{Appendix B: Choice of Membership Inference Attack}
\label{ap:b}
\begin{figure}[htbp]
    \centering
    \begin{subfigure}[b]{0.5\textwidth}
        \centering
        \begin{tikzpicture}
            \begin{axis}[
                width=\textwidth, height=3cm,
                xlabel={top-k\%},
                ylabel={Failure Rate},
                ymax=0.05,
                ymin=0,
                grid=both,
               legend style={at={(0.5,1.3)}, anchor=north,legend columns=-1,font=\scriptsize},
                ybar, 
                xtick=data, 
                bar width=0.1cm, 
                enlarge x limits=0.15, 
                xticklabels={5,10,15,20,25,30}, 
                symbolic x coords={0.05,0.1,0.15,0.2,0.25,0.3}, 
                xticklabels={5,10,15,20,25,30}
            ]

            \addplot[color=black, fill=black,draw=none, opacity=1] table[x=k, y expr=1-\thisrow{A}, col sep=comma] {q2_exact_cifar10_largest_attack.csv};
            \addlegendentry{A-LiRA}
            \addplot[color=black, fill=black,draw=none, opacity=0.6] table[x=k, y expr=1-\thisrow{online}, col sep=comma] {q2_exact_cifar10_largest_attack.csv};
            \addlegendentry{Online-LiRA}
            \addplot[color=black, fill=black,draw=none, opacity=0.3] table[x=k, y expr=1-\thisrow{offline}, col sep=comma] {q2_exact_cifar10_largest_attack.csv};
            \addlegendentry{Offline-LiRA}
            \end{axis}
        \end{tikzpicture}
        \caption{Unlearn top-k\%-risk samples of CIFAR-10.}
        \label{fig:top-down-exact-q2-attack}
    \end{subfigure}%
    \hfill
    \begin{subfigure}[b]{0.5\textwidth}
        \centering
        \begin{tikzpicture}
            \begin{axis}[
                width=\textwidth, height=3cm,
                xlabel={top-k\%},
                ylabel={Failure Rate},
                ymax=0.05,
                ymin=0,
                grid=both,
               legend style={at={(0.5,1.3)}, anchor=north,legend columns=-1,font=\scriptsize},
                ybar, 
                xtick=data, 
                bar width=0.1cm, 
                enlarge x limits=0.15, 
                xticklabels={5,10,15,20,25,30}, 
                symbolic x coords={0.05,0.1,0.15,0.2,0.25,0.3}, 
                xticklabels={5,10,15,20,25,30}
            ]

            \addplot[color=black, fill=black, draw=none,opacity=1] table[x=k, y expr=1-\thisrow{A}, col sep=comma] {q2_exact_cifar10_smallest_attack.csv};
            \addlegendentry{A-LiRA}
            \addplot[color=black, fill=black, draw=none,opacity=0.6] table[x=k, y expr=1-\thisrow{online}, col sep=comma] {q2_exact_cifar10_smallest_attack.csv};
            \addlegendentry{Online-LiRA}
             \addplot[color=black, fill=black, draw=none,opacity=0.3] table[x=k, y expr=1-\thisrow{offline}, col sep=comma] {q2_exact_cifar10_smallest_attack.csv};
            \addlegendentry{Offline-LiRA}
            \end{axis}
        \end{tikzpicture}
        \caption{Unlearn bottom-k\%-risk samples of CIFAR-10.}
        \label{fig:bottom-up-exact-q2-attack}
    \end{subfigure}
    \label{fig:unlearning_trends}
    \caption{Criterion 2: A-LiRA and online-LiRA detect more samples that failed to meet Criterion 2.}
    \label{fig:exact-q2-attack}
\end{figure}
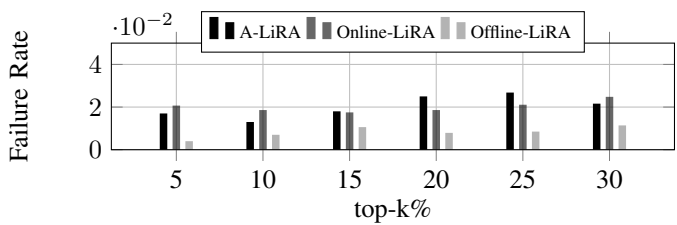
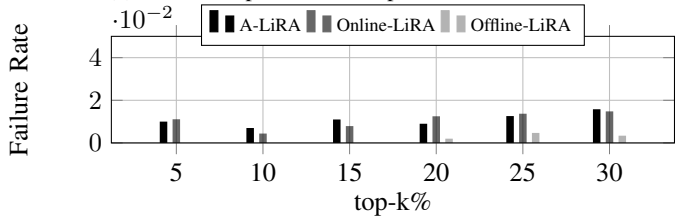

In this section, we analyze how the choice of attack affects the estimation of privacy risks. Intuitively, in a worst-case scenario, more powerful attacks provide a better estimate of a sample's privacy risk, as stronger attacks are more likely to detect samples that fail to meet both criteria. We evaluate online-LiRA, offline-LiRA, and A-LiRA by generating $\ln\left(\text{TPR}/\text{FPR}\right)$ in this experiments. To exclude the impact of the unlearning samples, we do not use approximate unlearning but directly exclude the unlearning samples and retrain the model. Since the unlearned samples are not used to train the model at all, we only evaluate Criterion 2 in this experiment. As shown in Figure~\ref{fig:exact-q2-attack}, online-LiRA and A-LiRA perform similarly, while offline-LiRA performs worse. However, as noted in Table~\ref{tab:attacking}, offline-LiRA is more efficient than both online-LiRA and A-LiRA, making it a viable alternative when efficiency is the primary concern for model builders.

\end{document}